\documentclass[nohyperref]{article}

\usepackage{microtype}
\usepackage{graphicx}
\usepackage{subcaption}
\usepackage{booktabs} 
\usepackage{multirow}
\usepackage{enumitem}

\usepackage{hyperref}

\usepackage[accepted]{icml2023}

\usepackage{amsmath}
\usepackage{amssymb}
\usepackage{mathtools}
\usepackage{amsthm}

\usepackage{amsfonts,bbm}
\newcommand{\R}{\mathbb{R}}

\newcommand{\1}{\mathbbm{1}}

\DeclareMathOperator*{\argmin}{arg\,min}
\DeclareMathOperator*{\argsort}{arg\,sort}

\usepackage[capitalize,noabbrev]{cleveref}

\theoremstyle{plain}

\theoremstyle{definition}

\theoremstyle{remark}

\usepackage[textsize=tiny]{todonotes}

\icmltitlerunning{Which Tricks Are Important for Learning to Rank?}

\begin{document}

\twocolumn[
\icmltitle{Which Tricks Are Important for Learning to Rank?}



\icmlsetsymbol{equal}{$\dagger$}

\begin{icmlauthorlist}
\icmlauthor{Ivan Lyzhin}{yandex}
\icmlauthor{Aleksei Ustimenko}{share,equal}
\icmlauthor{Andrey Gulin}{yandex}
\icmlauthor{Liudmila Prokhorenkova}{research}
\end{icmlauthorlist}

\icmlaffiliation{yandex}{Yandex, Moscow, Russia}
\icmlaffiliation{share}{ShareChat, London, UK}
\icmlaffiliation{research}{Yandex Research, Amsterdam, The Netherlands}

\icmlcorrespondingauthor{Aleksei Ustimenko}{research@aleksei.uk}
\icmlcorrespondingauthor{Liudmila Prokhorenkova}{ostroumova-la@yandex-team.ru}

\icmlkeywords{Learning to rank, GBDT, LambdaMART, StochasticRank, YetiRank, YetiLoss}

\vskip 0.3in
]



\printAffiliationsAndNotice{$^\dagger$The work was done while working at Yandex.}  

\begin{abstract}
Nowadays, state-of-the-art learning-to-rank methods are based on gradient-boosted decision trees (GBDT). The most well-known algorithm is Lambda\-MART which was proposed more than a decade ago. Recently, several other GBDT-based ranking algorithms were proposed. In this paper, we thoroughly analyze these methods in a unified setup. In particular, we address the following questions. Is direct optimization of a smoothed ranking loss preferable over optimizing a convex surrogate? How to properly construct and smooth surrogate ranking losses? To address these questions, we compare LambdaMART with YetiRank and Stochastic\-Rank methods and their modifications. We also propose a simple improvement of the YetiRank approach that allows for optimizing specific ranking loss functions. As a result, we gain insights into learning-to-rank techniques and obtain a new state-of-the-art algorithm.
\end{abstract}

\section{Introduction}\label{sec:introduction}

\emph{Learning to rank} (LTR) is a central problem in information retrieval~\cite{burges2010ranknet}. The objective of LTR is to rank a given set of items to optimize the overall utility of the list. In information retrieval, given a query and a list of candidate documents, it is desirable to rank the documents in decreasing order of \emph{relevance} to the query. Usually, LTR methods learn a function computing \emph{scores} for all documents and then sort the documents according to their scores. Thus, a major challenge for LTR is that the sorting operation is non-differentiable, which prevents effective gradient-based optimization.

Despite the recent success of neural approaches in various machine learning tasks, gradient-boosted decision trees (GBDT) are still state-of-the-art algorithms for \emph{tabular} datasets containing heterogeneous and noisy features~\cite{gorishniy2021revisiting,katzir2021net}. In particular, GBDT methods outperform other approaches for the learning-to-rank problem. A recent paper by~\citet{qin2021neural} observes that for standard tabular LTR datasets, the best results are achieved by a classic LambdaMART algorithm~\cite{wu2010adapting} proposed more than a decade ago and implemented within the LightGBM library~\cite{LightGBM}. Thus, our paper focuses on GBDT-based learning-to-rank methods and provides an extensive study of existing and new approaches and their important algorithmic details.

In addition to LambdaMART, we revisit another long-known algorithm called YetiRank~\cite{gulin2011winning}. Yeti\-Rank won The Transfer Learning Track of Yahoo!'s LTR challenge in 2010 but was not much explored after that. The main algorithmic differences of YetiRank compared to LambdaMART are stochastic smoothing used during training to randomize the orderings and a different way to construct an upper bound for the loss function with a tighter approximation. Our empirical evaluation shows that Yeti\-Rank outperforms LambdaMART and other competitors in most cases. In addition, we show that YetiRank implicitly optimizes a particular form of the DCG ranking quality function. This observation allows us to modify the YetiRank procedure by re-weighting the gradients according to a proper loss. The proposed modification YetiLoss further improves the performance of YetiRank for specific losses, for instance, $\mathrm{MRR}$ and $\mathrm{MAP}$. For $\mathrm{MAP}$, we obtain new state-of-the-art results with YetiLoss. We also explain why YetiLoss is expected to have more stable optimization and better generalization.

A recent work by~\citet{ustimenko2020stochasticrank} shows that the LambdaMART approach, implemented within the CatBoost library~\cite{catboost}, can be outperformed by an algorithm called StochasticRank. Stochastic\-Rank directly optimizes any given ranking loss with provable convergence guarantees. For this purpose, it uses stochastic smoothing of the loss function (without upper bounding). The obtained objective is often non-convex, so it is optimized with Stochastic Gradient Langevin Boosting~\cite{ustimenko2020stochasticrank}~--- a boosting algorithm designed to optimize non-convex functions. 

The main difference between StochasticRank and YetiLoss is that the former optimizes the training loss directly, while the latter constructs special convex bounds on the loss function at each iteration. While minimizing such bounds cannot guarantee reaching the global optima, convexity allows for more efficient optimization, which can often be beneficial. We provide extensive experiments to validate in which scenarios direct optimization is helpful.

To sum up, our main contributions are the following.
\begin{itemize}
    \item We conduct a thorough analysis of several state-of-the-art GBDT-based LTR algorithms. We provide a theoretical explanation of their differences and extensive empirical evaluation.
    \item We analyze the effect of several algorithmic details: optimizing the loss function directly versus constructing a convex upper bound, adding stochastic smoothing to the loss function, and different ways of constructing an upper bound. 
    \item We propose a simple extension of the YetiRank algorithm that handles arbitrary loss functions. The obtained algorithm achieves new state-of-the-art results for such loss functions as $\mathrm{MRR}$ and $\mathrm{MAP}$ that are often overlooked in LTR literature.
\end{itemize}

In the next section, we give the necessary background: define the learning-to-rank problem and popular quality functions, and then describe several well-known ranking algorithms. In Section~\ref{sec:YetiLoss}, we propose our modification of Yeti\-Rank that allows for optimizing any given ranking loss and also analyze and compare the properties of all the considered algorithms. In Section~\ref{sec:comparison}, we conduct a thorough comparison of existing LTR algorithms on several benchmarks, show that YetiLoss outperforms the competitors for specific ranking quality functions, and analyze the effect of the main algorithmic details on the quality of LTR. Section~\ref{sec:conclusion} concludes the paper.

\section{Background and Related Work}

\subsection{Learning-to-rank Problem}\label{sec:LTR_problem}

Learning to rank is a classic information retrieval problem~\cite{Liu:2009}. To formally define the LTR task, consider a query $q$ sampled from a distribution $Q$. For this query, we are given a set of documents $\{x_1, \ldots, x_{n_q}\}$ and a relevance vector $r = (r_1, \ldots, r_{n_q}) \in \mathbb{\R}^{n_q}$. Each document $x_i$ is represented as a vector of features that describes the query-document pair. 

A learning-to-rank model is a function $f(x)$ that takes the document's features $x_i$ and returns a value $z_i$ that is the predicted relevance for this document. For a query $q$, we compute the vector $z = (f(x_1), \ldots, f(x_{n_q})) \in \mathbb{R}^{n_q}$ called \emph{a vector of scores} and then compute $s = \mathrm{arg\,sort}(z) \in \mathbb{R}^{n_q}$ that is the ranking of documents predicted by the model~$f(x)$.

A ranking loss function $L(z,r)$ is a function that measures how well the ranking $s = \mathrm{arg\,sort}(z)$ agrees with the relevance vector $r$. To simplify the notation, we also write $L(f, q)$ meaning that $\{x_1, \ldots, x_{n_q}\}$ and $r$ are contained in $q$ and $z = (f(x_1), \ldots, f(x_{n_q}))$ is computed internally.

The ultimate goal of LTR is to find a model $f(x) \in \mathcal{F}$ ($\mathcal{F}$ is a predefined class of models) such that:
\begin{equation}\label{eq:goal}
    f = \argmin_{f \in \mathcal{F}} \mathbb{E}_{q \sim Q} L(f, q)\,.
\end{equation}
Unfortunately, in practice, we do not know the distribution $Q$ but rather have an i.i.d.~sample $q_1, \ldots, q_N \sim Q$ which we call the \emph{training} dataset. Thus, instead of~\eqref{eq:goal}, we solve:
\begin{equation}\label{eq:empirical}
    f = \argmin_{f \in \mathcal{F}} \frac{1}{N}\sum_{i=1}^N L(f, q_i)\,.
\end{equation}
The function $\mathcal{L}_N(f) := \frac{1}{N}\sum_{i=1}^N L(f, q_i)$ is called an \emph{empirical loss function}.

The main difficulty of the LTR problem is the fact that the function $\mathcal{L}_N(f)$ is locally constant as a function of $z$, discontinuous, and non-convex in $z$, since each $L(z, r)$ is defined by a permutation $s = \argsort(z)$. As a result, $\mathcal{L}_N(f)$ does not have a well-defined derivative (derivatives of $L(z, r)$ are everywhere zero due to a local constancy). Thus, one cannot directly optimize this loss with standard methods like gradient boosting or backpropagation in neural networks.

\subsection{Ranking Loss Functions}\label{sec:losses}

In this section, we define some widely used ranking loss functions. First, recall that all ranking losses depend on the permutation $s = \mathrm{arg\,sort}(z)$. However, it can be the case that two documents $x_i$ and $x_j$ have the same score, i.e., $z_i = z_j$. This situation is called a \emph{tie}, and in this case, $\mathrm{arg\,sort}(z)$ is not well-defined. Ties may occur due to the model's discrete structure (e.g., if $f$ is a GBDT model) or when two different documents have the same feature vectors. Although ties are often neglected in the LTR literature, they can affect the optimization procedure and lead to an overestimated ranking quality~\cite{ustimenko2020stochasticrank}. Therefore, \citet{ustimenko2020stochasticrank} propose resolving the ties as $s = \mathrm{arg\,sort}((z, -r))$, meaning that the documents are ordered according to $z$, but in the case of ties, we order them accordingly to $-r$, i.e., put the less relevant document first. In other words, the \emph{worst} permutation is used for ties. In this paper, we follow this tie-resolution policy.

Let $s = \mathrm{arg\,sort}((z, -r))$ and let $n = n_q$ be the length of the vectors $z$ and $r$. Probably the most well-known ranking quality function is $\mathrm{DCG}@k$:
\begin{equation}\label{eq:dcg}
    \mathrm{DCG}@k(z, r) = \sum_{i=1}^{\min\{n, k\}} \frac{2^{r_{s_i}} - 1}{2^4\log_2 (i + 1)}\,, 
\end{equation}
where $r_i\in [0, 4]$ are relevance labels. This quality function is called \emph{Discounted Cumulative Gain}: we divide the \textit{gain} for the relevance by the \textit{discount} for a lower position. Different options for \emph{gain} and \emph{discount} functions are possible, and~\eqref{eq:dcg} is the most widely used combination.

$\mathrm{NDCG}@k$ is a normalized variant of $\mathrm{DCG}@k$:
\begin{equation*}
    \mathrm{NDCG}@k(z, r) = \frac{\mathrm{DCG}@k(z, r)}{\max_{z' \in \mathbb{R}^n}{\mathrm{DCG}@k}(z', r)}\,.
\end{equation*}
Usually, LTR papers report the values of $\mathrm{NDCG}@k$, while other measures can be of interest in practice. For instance, for binary relevance labels $r_j \in \{0, 1\}$, one can be interested in \emph{Average Precision} ($\mathrm{AP}$):
\begin{equation*}
    \mathrm{AP}(z,r) = \frac{ \sum_{i=1}^n \mathrm{P}(i) \, r_{s_i} } {\sum_{i=1}^n r_i}\,, \,\text{ where } \mathrm{P}(i) = \frac 1 i \sum_{j = 1}^i r_{s_j}\,.
\end{equation*}
\emph{Mean Average Precision} ($\mathrm{MAP}$) is $\mathrm{AP}$ averaged over the queries. Another measure used for binary labels is \textit{Reciprocal Rank} (called $\mathrm{MRR}$ after the averaging):
\begin{equation*}
    \mathrm{RR}(z, r) = \sum_{i=1}^{n} \frac{r_{s_i}}{i}\prod_{j=1}^{i-1} (1-r_{s_j}), \quad r_j \in \{0, 1\}\,,
\end{equation*}
which is the inverse rank of the first relevant document. Finally, {\it Expected Reciprocal Rank} ($\mathrm{ERR}$) is a similar measure applied to non-binary labels:
\begin{equation*}
    \mathrm{ERR}(z, r) = \sum_{i=1}^{n} \frac{r_{s_i}}{i}\prod_{j=1}^{i-1} (1-r_{s_j})\,, \quad r_j \in [0, 1]\,. 
\end{equation*}
The above examples are ranking \emph{quality} functions, i.e., higher values are preferred. When we need to convert a quality function $M(z,r)$ to a loss function, we can just consider $L(z,r) := 1 - M(z,r)$. Note that in our experiments, we report ranking quality values multiplied by 100. 

\subsection{Ranking Algorithms}\label{sec:algorithms}

This section describes LTR algorithms that are of particular interest to the current research. For more methods and a broader overview, we refer to~\citet{Liu:2009,ustimenko2020stochasticrank,qin2021neural}. 

As discussed in Section~\ref{sec:LTR_problem}, the ranking loss function $\mathcal{L}_N(f)$ is non-differentiable. Thus, if we want to apply a convenient optimization method like gradient boosting or backpropagation, we need to build a differentiable surrogate. One of the most popular approaches is to use the Majorization-Minimization technique, when at each iteration $t$ we build a function $l_t(z, r)$ that is usually convex and differentiable such that:
\begin{equation}\label{eq:upper_bound}
    L(z, r) \le l_t(z, r) \,\,\,\, \forall\, z \in \mathbb{R}^{n_q}\,.
\end{equation}
After that, we sum up the upper bounds~\eqref{eq:upper_bound} for all the queries and obtain:\footnote{As before, we denote by $l_t(f, q)$ the values of $l_t(z, r)$, where we set $z = (f(x_1), \ldots, f(x_{n_q}))$.}
\begin{equation}\label{eq:L*}
    \mathcal{L}_N(f) \le \frac{1}{N}\sum_{i=1}^N l_t(f, q_i) = \mathcal{L}_t^*(f)\,.
\end{equation}
The loss function $\mathcal{L}_t^*(f)$ can be optimized directly by gradient boosting or backpropagation.

Many popular learning-to-rank algorithms like Lambda\-Rank~\cite{quoc2007learning}, LambdaMART~\cite{wu2010adapting}, or Lambda\-Loss~\cite{lambdaloss} fall into this category.
In this section, we mostly follow the reasoning of~\citet{lambdaloss}. 

To get a differentiable upper bound, these algorithms first build a bound of the form:
\begin{equation}\label{eq:general-pair-upper}
    L(z, r) \le \mathrm{const} + \sum_{i, j=1}^{n_q} w_{ij}\cdot \mathbbm{1}_{z_j - z_i > 0} \,.
\end{equation}
After that, one can use the inequality
$\mathbbm{1}_{z_j - z_i > 0} \le \log_2 \left(1 + e^{-(z_i - z_j)}\right)$ to obtain a convex differentiable bound:
\begin{equation}
    L(z, r) \le \mathrm{const} + \sum_{i, j=1}^{n_q} w_{ij} \log_2 \left(1 + e^{-(z_i - z_j)}\right)\,.
\end{equation}

To be more precise, for $\mathrm{NDCG}@k$ LambdaRank and Lambda\-MART select the weights $w_{ij}$ as:
\begin{equation}\label{eq:weights}
    w_{ij} = \Delta_{ij}\mathrm{NDCG}@k(z, r)\cdot\mathbbm{1}_{r_i>r_j}\,,
\end{equation}
where $\Delta_{ij}\mathrm{NDCG}@k(z, r)$ denotes the difference in $\mathrm{NDCG}@k(z, r)$ caused by permuting $i$-th and $j$-th documents in the sorting defined by $z$.
Formally, recall that $s = \mathrm{arg\,sort}(z)$ and consider the inverse permutation $p = s^{-1}$. The value $p_i$ is the position of the $i$-th document in the list ordered by $z$. Then,
\begin{multline}\label{eq:delta_ndcg}
\Delta_{ij}\mathrm{NDCG}@k(z, r) = \frac{(2^{r_i} - 2^{r_j})}{2^4 \max_{z'} \mathrm{DCG}@k(z', r)} \\ \cdot \bigg|\frac{1}{\log_2 (1 + p_j)} - \frac{1}{\log_2(1 + p_i)}\bigg|\,.
\end{multline}

As proven by~\citet{lambdaloss}, using such $w_{ij}$ indeed gives a desired upper bound.
Similarly, one can adapt LambdaMART to any ranking metric by substituting $\Delta_{ij}\mathrm{NDCG}@k(z, r)$ in~\eqref{eq:weights} by the difference of the target loss function. It is important to note that the \emph{upper bound property may not hold} in this case.
However, empirical evidence shows that this trick works well in practice.

Now let us describe the YetiRank algorithm. The original formulation given by~\citet{gulin2011winning} does not specify which ranking metric is optimized, but the pairwise loss is similar to the ones described above. YetiRank uses the following convex differentiable loss:
\begin{equation}\label{eq:yeti-loss}
l_t(z, r) = - \sum_{i, j=1}^{n_q} w_{ij} \log \left(1 + e^{-(z_i - z_j)}\right)\,,
\end{equation}
where the weights are chosen as:
\begin{equation}\label{eq:yeti-weights}
w_{ij} = (r_i - r_j) \cdot \mathbbm{1}_{r_i>r_j} \cdot \mathbbm{E}_{\epsilon}[b^{\,p_i - 1} \cdot   \mathbbm{1}_{|p_i - p_j|=1} ]\,,
\end{equation}
where $b \in (0, 1)$ is a hyper-parameter and $p_i$ is a random permutation obtained as the inverse of $\mathrm{arg\,sort}(z + \epsilon)$ with $\epsilon = \log \frac{u}{1 - u}, u \sim U([0, 1]^{n_q})$, i.e., $\epsilon$ is distributed according to the Logistic distribution. In practice, to estimate $w_{ij}$, one adds random noise to the current scores to generate several random permutations. By default, ten permutations are used.\footnote{The complexity of computing the gradients is thus increased with the number of permutations, but this does not affect the overall time complexity much since the most time-consuming operation is building a decision tree, and this procedure is not affected.}
Importantly, for YetiRank, we can obtain only stochastic gradient estimates of $\nabla_z \, l_t(z, r)$ due to the randomness of $\epsilon$. We provide a deeper analysis of YetiRank in Section~\ref{sec:YetiLoss}.

The CatBoost library also has a modification of YetiRank called YetiRankPairwise~\citep{CatBoost_documentation}, which is described in~\citet{gulin2011winning}. It is a GBDT-specific modification of YetiRank, allowing one to get more accurate results on large datasets. However, in our preliminary experiments, we have not observed consistent improvements of Yeti\-Rank\-Pairwise over YetiRank. Since YetiRankPairwise is significantly slower than all other algorithms considered in this paper, we do not include it in our empirical analysis.

Finally, we overview the recently proposed Stochastic\-Rank algorithm~\cite{ustimenko2020stochasticrank}. In sharp contrast with the previous approaches that build convex differentiable surrogates to optimize a given non-convex ranking loss, StochasticRank smooths $L(z, r)$ directly using the Gaussian distribution $\mathcal{N}(-\mu r, \sigma I_{n_q})$:
\begin{equation}
    l(z, r) = \mathbbm{E}_{\epsilon \sim \mathcal{N}(0, I_{n_q})} L(z - \mu r + \sigma \epsilon, r)\,,
\end{equation}
where $\mu \ge 0$ and $\sigma > 0$ are hyper-parameters. Such smoothing leads to the loss function $\mathcal{L}_N(f, \sigma, \mu) = \frac{1}{N} \sum_{i=1}^N l(f, q_i)$ which minimum point is arbitrary close to the minimum of $\mathcal{L}_N(f)$, see~\citet{ustimenko2020stochasticrank} for the details. Moreover, the paper proposes novel stochastic gradient estimates to estimate the gradients more effectively compared to the log-derivative trick~\cite{Nesterov2017RandomGM}. Finally, StocasticRank relies on Stochastic Gradient Langevin Boosting (SGLB)~\cite{SGLB} that guarantees that the optimization would eventually reach the global optimum even for the non-convex loss function $\mathcal{L}_N(f, \mu, \sigma)$.

\section{Analysis of LTR Algorithms}\label{sec:YetiLoss}

In Section~\ref{sec:comparison}, we will show that YetiRank achieves state-of-the-art results for most datasets and quality functions. However, it is not specifically optimized for any of these metrics. In this section, we first discuss which loss function is optimized by YetiRank and then propose a simple modification called YetiLoss that can optimize any given ranking quality function. Finally, we discuss how YetiLoss relates to LambdaMART and StochasticRank.

\subsection{YetiLoss Algorithm}

Let us consider the following ranking quality function for a fixed $b \in (0, 1)$:
\begin{equation}\label{eq:ExpDCG}
    \mathrm{ExpDCG}(z, r) = \sum_{i=1}^{n_q} b^{i-1} r_{s_i}\,,
\end{equation}
where $s = \mathrm{arg\,sort}(z)$. This loss resembles the classic $\mathrm{DCG}$ given in~\eqref{eq:dcg}, but the gain is $r_{s_i}$ instead of $2^{r_{s_i}}-1$, while the discount is exponential ($b^{i-1}$) instead of logarithmic ($\log_2(i+1)$).

It follows from~\eqref{eq:yeti-weights} and~\eqref{eq:ExpDCG} that up to a constant multiplier, we have
\begin{equation*}
    w_{ij} = \mathbbm{1}_{r_i>r_j} \cdot\mathbbm{E}_{\epsilon} [\Delta_{ij}\mathrm{ExpDCG}(z + \epsilon, r) \cdot \mathbbm{1}_{|p_i - p_j|=1}]\,,
\end{equation*}
which means that YetiRank optimizes $\mathrm{ExpDCG}$ but in a smoothed form.

This observation suggests that to adapt YetiRank for an arbitrary loss function, we need to replace $\Delta_{ij}\mathrm{ExpDCG}$ by the difference in the desired ranking loss. For $\mathrm{NDCG}@k$, the expression is given in~\eqref{eq:delta_ndcg}. For $\mathrm{MRR}$, we have:
\begin{equation*}\label{eq:delta_mrr}
\Delta_{ij}\mathrm{MRR}(z, r)
= \bigg| \frac{1}{p_i} - \frac{1}{p_j} \bigg| \cdot \prod_{k=1}^{\min\{i,j\}-1} (1-r_{s_j}) \cdot \1_{r_i = 1 - r_j}.
\end{equation*}
Analogously, we obtain the expressions for $\mathrm{ERR}$ and $\mathrm{MAP}$. Similarly to LambdaMART, for an arbitrary metric (rather than $\mathrm{NDCG}$), we do not have the upper bound guarantees. However, our experiments show that YetiLoss outperforms standard YetiRank for the $\mathrm{MRR}$ and $\mathrm{MAP}$ and in half of the cases for $\mathrm{ERR}$.

\subsection{Comparison of YetiLoss and LambdaMART}

For a given ranking loss function $M(z, r)$, LambdaMART and YetiLoss use the following weights:
\[
w_{ij}^{LM} = \Delta_{ij}M(z, r)\cdot\mathbbm{1}_{r_i>r_j}\,,
\]
\[
w_{ij}^{YL} = \mathbbm{E}_{\epsilon} \Delta_{ij}M(z + \epsilon, r) \cdot \mathbbm{1}_{r_i>r_j} \cdot \mathbbm{1}_{|p_i - p_j|=1} \,.
\]
Even though both weights share the term $\Delta_{ij}M \cdot\mathbbm{1}_{r_i>r_j}$, they have principal differences, which we discuss below.

First, let us note that $w^{LM}$ is a discontinuous function of the argument $z$ (model predictions). Hence, it may drastically change when the model $f$ changes only slightly. In contrast, $w^{YL}$ is a smooth function of $z$ due to smoothing with an absolutely continuous distribution. Training with a smooth objective is expected to be more robust; thus, we expect YetiLoss to have a better generalization.

Let us also note that a similar smoothing is proposed in a recent paper by~\citet{48689}. The authors use a different smoothing distribution and also add noise into $f_i - f_j$ that appears in the logarithmic term that is multiplied by the weights. The authors confirm that smoothing the loss function improves the performance. While comparing with~\citet{48689} is out of the scope of the current paper since we focus on widely used open-source implementations, we do empirically evaluate the importance of smoothing and different smoothing distributions in Section~\ref{sec:details}.

The second difference is that $w^{YL}$ contains $\Delta_{ij}$ only for the neighboring documents with $|p_i - p_j| = 1$. In other words, YetiLoss uses only those pairwise permutations that \emph{can be realized} by small changes in $z$. In contrast, $w^{LM}$ does not have this restriction and contains the pairwise permutations that \emph{cannot be realized} by small changes of $z$. This implies that for LambdaMART, the resulting gradients may contain redundant information about the permutations that cannot be achieved within one gradient-boosting step with a small step size. Hence, the obtained upper bound on the ranking loss can be too loose. In Section~\ref{sec:details}, we empirically evaluate the effect of this difference.

\subsection{Comparison of YetiLoss and StochasticRank}

The approach underlying the StochasticRank algorithm significantly differs from both YetiLoss and LambdaMART (see Section~\ref{sec:algorithms}). However, it turns out that Stochastic\-Rank has a structural similarity with YetiLoss but not with Lambda\-MART.
This similarity lies in pairwise permutations that YetiLoss incorporates. 
We already mentioned that LambdaMART relies on all possible pairwise permutations to build the surrogate loss $\mathcal{L}_t^*$. In contrast, YetiLoss uses only those that permute the neighborhood elements in the sorted list.
Similarly, we can show that StochasticRank also relies only on such permutations. 
To see that, we write down the formula for $\frac{\partial}{\partial z_i} l(z, r, \sigma, \mu)$ estimate that is introduced by~\citet{ustimenko2020stochasticrank} and is called Coordinate Conditional Sampling (CCS). Let us assume for simplicity that $\mu = 0$:
\begin{multline*}
\frac{\partial^{CCS}}{\partial z_i} l(z, r, \sigma) = (2\pi \sigma^2)^{-\frac{1}{2}} \\ \cdot \sum_{j=0}^{n_q} \delta_{i,j} L(z + \epsilon, r) e^{-\frac{1}{2\sigma^2}(z_i - z_j + \sigma(\epsilon_i - \epsilon_j))^2}\,,
\end{multline*}
where $\delta_{ij}$ corresponds to the difference of the ranking loss if we change only $z_i$ without changing the remaining values, so that the $i$-th item is placed after the $j$-th item for $j > 0$. For $j = 0$, it means the difference if we put the $i$-th item on the first position.

Next, if we assume that all $z_i$ are different, the properties of the super-exponential decay of $e^{-y^2}$ allow us to ensure that for small enough $\sigma \approx 0$ the latter can be approximated as:
\begin{multline*}
\frac{\partial^{CCS}}{\partial z_i} l(z, r, \sigma) \\ \approx (2\pi \sigma^2)^{-\frac{1}{2}} \cdot \left(\delta_{i,a} L(z + \epsilon, r) e^{-\frac{1}{2\sigma^2}(z_i - z_a)^2} \right. \\ \left. + \delta_{i,b} L(z + \epsilon, r) e^{-\frac{1}{2\sigma^2}(z_i - z_b)^2}\right),
\end{multline*}
where $a$ is the index of the largest $z_a$ that is smaller than $z_i$, and $b$ is the index of the smallest $z_b$ that is larger than $z_i$, i.e., \emph{that is exactly the pairwise permutation of neighboring items}.

We note that the approximation can be made arbitrarily precise by letting $\sigma \rightarrow 0_+$. We also note that $\sigma \approx 0$ is preferred by the algorithm's design to achieve the theoretical guarantees on the optimization of $\mathcal{L}_N$ by StochasticRank.

Such observation suggests that the pairwise permutations of neighboring items are the only permutations that can be achieved by small perturbations of the model we are learning. Hence, $\mathcal{L}_t^*$ that arises from YetiLoss contains more information about \emph{local} behavior of the ranking loss function $\mathcal{L}_N$.

The similarity between YetiLoss and StochasticRank allows us to address the following question in the next section: is direct optimization of the \emph{non-convex} smoothed loss function better than optimization of a \emph{convex} surrogate loss?

\section{Experiments}\label{sec:comparison}

This section empirically compares popular LTR algorithms, evaluates the proposed modification called YetiLoss, and analyzes the importance of several algorithmic details in a unified setup. 

\begin{table*}[t]
\caption{Learning-to-rank datasets}
\label{tab:datasets}
\vspace{5pt}
\begin{center}
\begin{tabular}{l|c|rrr|rrr}
\toprule
\multirow{2}{*}{Dataset} & \multirow{2}{*}{\# features} & \multicolumn{3}{c|}{ \# queries} & \multicolumn{3}{c}{ \# documents} \\
 & & train\, & validation & test\,\, & train\,\,\,\,\, & validation & test\,\,\,\,\, \\
\midrule
Web10K & 136 & 6,000 & 2,000 & 2,000 & 723,412 & 235,259 & 241,521 \\
Web30K & 136 & 18,919 & 6,306 & 6,306 & 2,270,296 & 747,218 & 753,611 \\
Yahoo S1 & 699 & 19,944 & 2,944 & 6,983 & 473,134 & 71,083 & 165,660 \\
Yahoo S2 & 699 & 1,266 & 1,266 & 3,798 & 34,815 & 34,881 & 103,174 \\
Istella & 220 & 20,307 & 2,912 & 9,799 & 5,497,064 & 1,828,561 & 3,129,004 \\
Istella-S & 220 & 19,246 & 7,211 & 6,562 & 2,043,304 & 684,076 & 681,250 \\
\bottomrule
\end{tabular}
\end{center}
\vskip -0.1in
\end{table*}

\subsection{Setup}\label{sec:setup}

\paragraph{Datasets} We use six publicly available datasets. The first two are Web10K and Web30K released by Microsoft~\cite{QinL13}. Following previous studies~\cite{qin2021neural,ustimenko2020stochasticrank,lambdaloss}, we use Fold~1 for these two datasets. We also use two datasets from YAHOO! Learning to Rank Challenge~\cite{Chapelle2010}. Finally, we take Istella and Istella-S datasets~\cite{dato2016fast}. All datasets except for Istella are pre-divided into the train, validation, and test sets. For Istella, there is no standard validation set, so we randomly divided the train part into train and validation. Table~\ref{tab:datasets} overviews the datasets used in the current study.

\paragraph{Ranking loss functions} 
We compare the algorithms according to all quality functions described in Section~\ref{sec:losses}. 

The first one is $\mathrm{NDCG}@10$, which is very popular in LTR research. In some of our experiments, we also use $\mathrm{NDCG}@1$ and $\mathrm{NDCG}@5$. We observe no significant differences in the obtained results, so we omit these losses from the main text. Some results for $\mathrm{NDCG}@1$ and $\mathrm{NDCG}@5$ can be found in Appendix.

The second quality function is $\mathrm{MRR}$, which is a well-known click-based metric. As $\mathrm{MRR}$ requires binary labels, we binarize each label as $\widetilde{y}_i := \1_{\{y_i > 0\}}$. $\mathrm{MRR}$ is often used in practice while it is much less studied than $\mathrm{NDCG}@k$. 

We also consider $\mathrm{MAP}$, for which we binarize the labels in the same way. 

Finally, for $\mathrm{ERR}$, we convert the labels to $[0,1]$ as $\widetilde{y}_i := y_i/4$. Using all these different loss functions is essential for analyzing the generalizability of the approaches to different ranking tasks.

\paragraph{Algorithms} 
All the algorithms are implemented and available within the official CatBoost gradient boosting library~\cite{catboost,CatBoost_documentation}.

For YetiRank, we use the original implementation, which we modify to obtain YetiLoss compatible with $\mathrm{NDCG}$, $\mathrm{MRR}$, $\mathrm{ERR}$, and $\mathrm{MAP}$. 

For StochasticRank, we use the official implementation for $\mathrm{NDCG}$, while we also add $\mathrm{MRR}$ and $\mathrm{ERR}$ loss functions to the optimization. MAP is not supported in StochasticRank out of the box, and \citet{ustimenko2020stochasticrank} only provide algorithms for efficiently computing ERR-like and DCG-like metrics. Making an efficient and correct gradient estimate for MAP in StochasticRank can be a subject of a separate investigation, which is why we only implemented $\mathrm{MRR}$ and $\mathrm{ERR}$.

For Lambda\-MART, we implement the original algorithm~\cite{wu2010adapting} within the CatBoost library with all ranking quality functions considered in our analysis. 

For completeness of the analysis, we also add the CatBoost trained in the QueryRMSE regime~\cite{CatBoost_documentation}. In this case, the algorithm optimizes RMSE averaged over the queries. This simple loss function can be considered as a baseline for other methods.

Finally, we also add the implementation of LambdaMART provided by Microsoft within the LightGBM library~\cite{LightGBM}. This implementation is known to outperform other open-source versions and to achieve state-of-the-art LTR results for $\mathrm{NDCG}$~\cite{qin2021neural}, while it does not allow for optimizing $\mathrm{MRR}$, $\mathrm{ERR}$, or $\mathrm{MAP}$. Let us note, however, that LightGBM and CatBoost are different implementations of GBDT. This fact prevents a direct comparison of LTR algorithms. For instance, in contrast to LightGBM, CatBoots uses oblivious decision trees, where the same splitting criterion is used for all tree nodes at a given depth.

\paragraph{Parameter tuning}
For parameter tuning, we use 20 iterations of random search followed by 80 iterations of Bayesian optimization~\cite{balandat2020botorch}.\footnote{Small variations of the results compared to the previous papers can be explained by a particular parameter optimization used.} For all algorithms, we set the maximum number of trees to 1000. We choose the best parameters, including the optimal number of trees, using the value of the desired loss function on the validation set. The list of tuned parameters is given in Appendix~\ref{app:setup}.

\begin{table*}[t]
\caption{Comparison of learning-to-rank algorithms with tuned hyper-parameters}
\label{tab:main_results}
\vspace{5pt}
\begin{center}
\begin{tabular}{lcccccccccc}
\toprule
Algorithm & Web10K & Web30K & Yahoo S1 & Yahoo S2 & Istella & Istella-S \\
\midrule
 & \multicolumn{6}{c}{$\mathrm{NDCG}@10$} \\
\midrule
LightGBM & 50.39 & 52.21 & 78.88 & 77.25 & 72.89 & 76.48 \\
QueryRMSE  & 50.55 & 52.19 & 79.10 & \textbf{78.01} & 69.66 & 74.83 \\
LambdaMART & 49.38 & 50.93 & 78.63 & 77.31 & 69.83 & 75.55 \\
StochasticRank & 50.34 & 51.63 & 78.98 & 77.77 & 68.46 & 74.88 \\
YetiRank & \textbf{50.75} & \textbf{52.31} & \textbf{79.19} & \textbf{77.99} & \textbf{73.02} & \textbf{77.15} \\
YetiLoss & \textbf{50.76} & 51.90 & 78.67 & 77.77 & 71.60 & 76.32 \\
\midrule
& \multicolumn{6}{c}{$\mathrm{MRR}$} \\
\midrule
QueryRMSE & 83.47 & 85.38 & 91.03 & 92.56 & 96.23 & 97.63 \\
LambdaMART & 82.59 & 84.12 & 90.75 & 92.34 & 95.18 & 96.96 \\
StochasticRank & 83.51 & 85.43 & 90.72 & 93.08 & 96.54 & 97.83 \\
YetiRank & \textbf{84.21} & 85.58 & \textbf{91.20} & 93.05 & 96.70 & 97.58 \\
YetiLoss & 84.17 & \textbf{86.01} & 90.88 & \textbf{93.21} & \textbf{97.16} & \textbf{97.97} \\
\midrule
& \multicolumn{6}{c}{$\mathrm{MAP}$} \\
\midrule
QueryRMSE & 62.11 & 63.51 & 86.06 & 88.40 & 78.11 & 88.59 \\
LambdaMART & 57.93 & 58.58 & 85.91 & 89.14 & 80.80 & 90.28 \\
YetiRank & 62.07 & 63.39 & 85.94 & 88.00 & 80.01 & 89.09 \\
YetiLoss & \textbf{62.62} & \textbf{64.03} & \textbf{86.46} & \textbf{89.37} & \textbf{82.66} & \textbf{91.07} \\
\midrule
& \multicolumn{6}{c}{$\mathrm{ERR}$} \\
\midrule
QueryRMSE & 57.27 & 59.30 & \textbf{66.76} & 67.16 & 85.29 & 85.55 \\
LambdaMART & 56.54 & 58.76 & 66.22 & 66.55 & 84.71 & 85.24 \\
StochasticRank & 56.94 & 59.34 & 66.64 & 67.17 & 85.44 & 85.98 \\
YetiRank & 57.41 & \textbf{59.53} & 66.69 & \textbf{67.27} & \textbf{87.29} & \textbf{86.97} \\
YetiLoss & \textbf{57.46} & \textbf{59.53} & 66.71 & 67.15 & 86.83 & 86.59 \\
\bottomrule
\end{tabular}
\end{center}
\vskip -0.1in
\end{table*}

\subsection{Comparison of LTR Algorithms}

The results of the comparison are present in Table~\ref{tab:main_results}. The best results are highlighted. The differences between the highlighted results and other algorithms are statistically significant according to the paired one-tailed t-test with p-value $< 0.05$. 

We make the following observations. First, we confirm the conclusion from~\citet{ustimenko2020stochasticrank} that StochastiRank outperforms LambdaMART in most cases. However, YetiRank beats both of them, leading to state-of-the-art results, especially for $\mathrm{NDCG}@10$. In turn, the proposed YetiLoss modification of YetiRank achieves the best results for $\mathrm{MAP}$ outperforming the competitors by a considerable margin. It is also the best on most datasets for $\mathrm{MRR}$ and performs similarly to YetiRank for $\mathrm{ERR}$. For $\mathrm{NDCG}@10$, YetiRank outperforms YetiLoss: this can be explained by the fact that the loss function optimized within YetiRank is similar to $\mathrm{NDCG}$.

\begin{table*}
\caption{The effect of stochastic smoothing for YetiLoss}
\label{sec:noise}
\vspace{5pt}
\begin{small}
\begin{center}
\begin{tabular}{l|ccc|ccc|ccc}
\toprule
 & \multicolumn{3}{c|}{$\mathrm{NDCG}$@10} & \multicolumn{3}{c|}{$\mathrm{MRR}$} & \multicolumn{3}{c}{$\mathrm{MAP}$} \\ 
 & Web10K & Yahoo S2 & Istella-S & Web10K & Yahoo S2 & Istella-S & Web10K & Yahoo S2 & Istella-S \\
\midrule
Logistic & 50.76 & \textbf{77.77} & \textbf{76.32} & 84.17 & 93.21 & \textbf{97.97} & \textbf{62.62} & \textbf{89.37} & 91.07 \\
Gaussian & \textbf{50.79} & 77.55 & 76.28 & \textbf{84.28} & \textbf{93.61} & 97.83 & 62.52 & \textbf{89.37} & \textbf{91.08} \\ 
No smoothing & 48.49 & 75.29 & 74.55 & 82.70 & 91.94 & 97.38 & 61.32 & 88.13 & 89.89 \\
\bottomrule
\end{tabular}
\end{center}
\end{small}
\end{table*}

\begin{figure*}
    \centering
    \begin{subfigure}{0.45\textwidth}
        \centering
        \includegraphics[width=\linewidth]{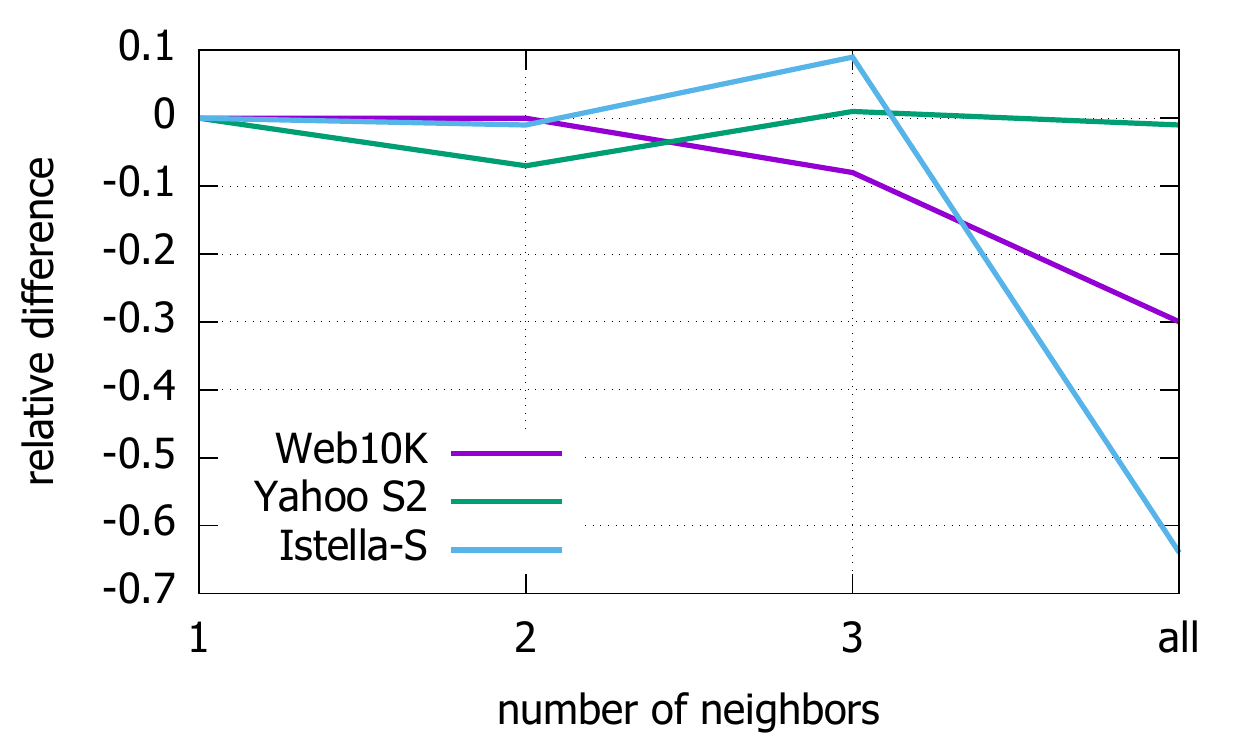}
        \caption{$\mathrm{MAP}$}
    \end{subfigure}
    \begin{subfigure}{0.45\textwidth}
        \centering
        \includegraphics[width=\linewidth]{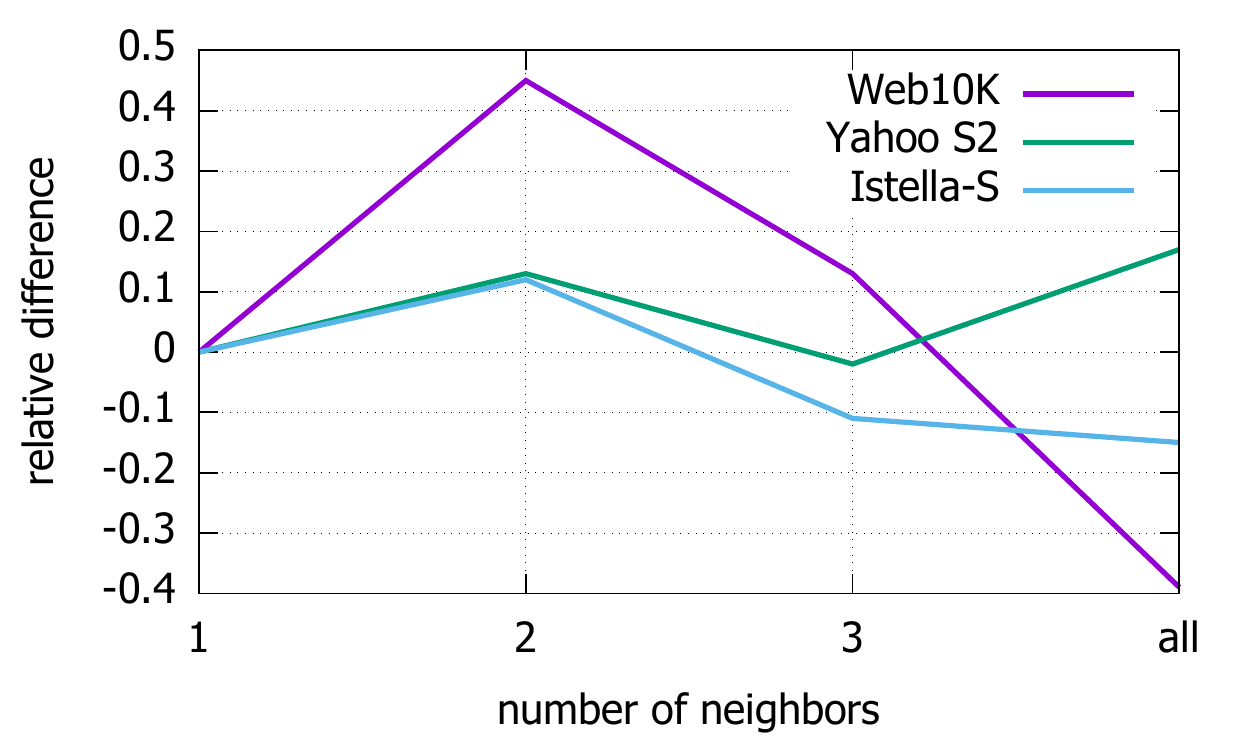}
        \caption{$\mathrm{ERR}$}
    \end{subfigure}
    \caption{The effect of the number of neighbors, relative loss compared to YetiLoss}
    \label{fig:neighbors}
\end{figure*}

Interestingly, the simple QueryRMSE method is very competitive in some cases. For instance, for $\mathrm{NDCG}@10$, it is superior to LightGBM, LambdaMART, and StochasticRank in at least half of the cases. QueryRMSE also achieves the best results on Yahoo S2 for $\mathrm{NDCG}@10$ and on Yahoo S1 for $\mathrm{ERR}$. However, in most cases, YetiLoss is significantly better than QueryRMSE. Our experiments with QueryRMSE show that considering such simple ranking-agnostic baselines is important for a fair analysis.

By comparing YetiLoss with StochasticRank, we can see that convexity of the optimization problem is usually preferred over direct optimization of a non-convex smoothed ranking loss: except for some results on Yahoo datasets, YetiLoss outperforms StochasticRank.

To better understand the differences between LTR algorithms, we also analyze their generalization ability. These results are shown in Appendix~\ref{app:experiments}.

\paragraph{Comparison with default hyper-parameters} To evaluate the effect of parameter tuning on the obtained experimental result, we additionally evaluate all the algorithms with their default set of hyper-parameters (only the optimal number of trees is chosen on the validation set). The results are shown in Table~\ref{tab:main_results_default}. Note that all our observations are also present for this setup. Indeed, YetiLoss is the best with a significant margin for $\mathrm{MRR}$ and $\mathrm{MAP}$. Moreover, it wins on half of the datasets for $\mathrm{ERR}$. For $\mathrm{NDCG}$@10, similarly to Table~\ref{tab:main_results}, the best results are achieved with YetiRank.

\begin{table*}[t]
\caption{Comparison of learning-to-rank algorithms with default hyper-parameters}
\label{tab:main_results_default}
\vspace{5pt}
\begin{center}
\begin{tabular}{lcccccccccc}
\toprule
Algorithm & Web10K & Web30K & Yahoo S1 & Yahoo S2 & Istella & Istella-S \\
\midrule
 & \multicolumn{6}{c}{$\mathrm{NDCG}@10$} \\
\midrule
LightGBM & 
\textbf{50.34} & 
\textbf{51.06} & 
78.34 & 
76.96 & 
67.71 & 
73.67 \\
QueryRMSE & 
49.82 & 
50.68 & 
78.34 & 
78.03 & 
65.45 & 
72.60 \\
LambdaMART & 
48.56 & 
49.50 & 
77.44 & 
76.46 & 
66.25 & 
73.37 \\
YetiRank & 
50.18 & 
50.91 & 
\textbf{78.58} & 
\textbf{78.15} & 
\textbf{68.86} & 
\textbf{74.72} \\
YetiLoss & 
50.14 & 
50.91 & 
78.03 & 
77.55 & 
68.73 & 
74.62 \\
\midrule
& \multicolumn{6}{c}{$\mathrm{MRR}$} \\
\midrule
QueryRMSE & 83.44 & 84.87 & 90.78 & 93.25 & 94.76 & 96.89 \\
LambdaMART & 82.36 & 83.97 & 90.72 & 91.84 & 94.73 & 96.81 \\
YetiRank & 	83.81 & 85.10 & 91.02 & 93.16 & 95.59 & 97.09 \\
YetiLoss & \textbf{84.50} & \textbf{85.98} & \textbf{91.14} & \textbf{93.76} & \textbf{96.40} & \textbf{97.71} \\
\midrule
& \multicolumn{6}{c}{$\mathrm{MAP}$} \\
\midrule
QueryRMSE & 61.85 & 63.00 & 85.66 & 88.40 & 72.61 & 85.95 \\
LambdaMART & 56.89 & 55.81 & 85.92 & 89.14 & 75.21 & 87.49 \\
YetiRank & 61.80 & 63.03 & 85.66 & 88.00 & 75.36 & 86.82 \\
YetiLoss & \textbf{62.43} & \textbf{63.58} & \textbf{86.41} & \textbf{89.36} & \textbf{76.61} & \textbf{88.14} \\
\midrule
& \multicolumn{6}{c}{$\mathrm{ERR}$} \\
\midrule
QueryRMSE & 56.98 & 58.64 & 66.26 & 67.20 & 83.06 & 84.14 \\
LambdaMART & 55.92 & 57.74 & 65.99 & 66.41 & 83.01 & 84.41 \\
YetiRank & 	
57.12  & 58.73 & \textbf{66.54} & \textbf{67.28} & 85.31 & \textbf{85.77} \\
YetiLoss & \textbf{57.16} & \textbf{58.80} & 66.50 & 67.22 & \textbf{85.46} & 85.71 \\
\bottomrule
\end{tabular}
\end{center}
\vskip -0.1in
\end{table*}

\subsection{Analysis of Algorithmic Details}\label{sec:details}

\paragraph{Stochastic treatment of scores}

First, we analyze the effect of stochastic smoothing of weights. Recall that the presence of such smoothing is the key difference between YetiLoss and LambdaMART. Table~\ref{sec:noise} confirms that smoothing is essential for Yeti\-Loss. On the other hand, a particular shape of smoothing is not so important: the results obtained with Gaussian and Logistic smoothing are similar.

\paragraph{Number of neighbors} Another difference between YetiLoss and LambdaMART is that YetiLoss uses only neighboring documents with $|p_i - p_j| = 1$ while computing the weights. To analyze the effect of this detail, we consider modifications of YetiLoss with $|p_i - p_j| = k$ for $k \in \{1, 2, 3\}$ (see Figure~\ref{fig:neighbors}). We also added the LambdaMART version, where all pairs are considered. We can see that there is no stable significant dependence on this parameter. However, choosing $k = 2$ is a good option. Note that smaller values of $k$ lead to more efficient algorithms.

\vspace{40pt}

\section{Conclusion and Discussion}\label{sec:conclusion}

This paper analyzes several GBDT-based state-of-the-art LTR algorithms and their modifications to determine which algorithmic details are important for better learning to rank. In our experiments, we found that YetiRank is currently the state-of-the-art LTR approach. We make a simple modification of YetiRank and propose an extension called YetiLoss that can optimize arbitrary ranking loss functions. The experiments confirm that YetiLoss is better at optimizing $\mathrm{MRR}$ and is considerably better than the competitors on all the datasets for $\mathrm{MAP}$. Another important contribution of our research is a thorough comparison of existing state-of-the-art learning-to-rank algorithms and their algorithmic details. 

Note that our work is limited to GBDT-based models. We chose this particular setup as GBDTs are known to outperform other approaches on classic tabular LRT benchmarks. Indeed, \citet{qin2021neural} compare the existing neural approaches with GBDT-based methods and show that if fairly evaluated, GBDTs still outperform neural methods.\footnote{The model proposed by~\citet{qin2021neural} is comparable with GBDTs, but it uses self-attention, so to score a particular document the neural network uses features of other documents that are not available to GBDT.} This agrees with the experiments of~\citet{gorishniy2021revisiting} comparing GBDTs and neural approaches on non-ranking tabular datasets. Importantly, GBDTs have other advantages useful in some applications: they are fast, easy to use, and do not require much parameter tuning. Also, in production scenarios (e.g., in search engines), neural networks usually compute some features that are then passed to a GBDT model to be combined with other heterogeneous features. GBDT models are often used on top of neural networks since they are known to be good at dealing with heterogeneous features of different natures and scales. That is why we strongly believe that the analysis of GBDT-based ranking is very important.

On the other hand, a useful direction for future research is to perform a similar systematic analysis for neural network models. All the discussed approaches (Yeti\-Loss, Query\-RMSE, and Stochastic\-Rank) can be straightforwardly applied to neural rankers. However, it is not guaranteed that the conclusions obtained in the current research would transfer to other types of models and problems, and a separate analysis is required, similarly to what is done for the Lambda\-Loss framework by~\citet{jagerman2022optimizing}. In addition, it would be useful to compare the neural rankers in a suitable setup like image retrieval or recommendation systems.

Finally, note that all the algorithms are implemented within the CatBoost open-source gradient boosting library for a fair comparison. We choose this particular library since it contains the original implementations of StochasticRank and YetiRank so that we can compare these methods and their modifications fairly. While we believe the obtained conclusions would transfer to other libraries, additional analysis is needed to confirm this.




\bibliography{ranking}
\bibliographystyle{icml2023}


\appendix

\newpage

\section{Experimental Setup}\label{app:setup}


We tune the following hyper-parameters:
\begin{itemize}[nolistsep]
    \item \emph{learning-rate}: log-uniform distribution over $[10^{-3}, 1]$ for all models;
    \item \emph{l2-leaf-reg}: $0$ for StochasticRank; log-uniform distribution over $[0.1, 100]$ for LightGBM; over $[0.001, 100]$ for other models;
    \item \emph{depth:} uniform distribution over $\{6, \ldots, 8\}$ for Yeti\-Rank and LambdaMART, over $\{6, \ldots, 10\}$ for StochasticRank, over $\{6, \ldots, 16\}$ for LightGBM;
    \item \emph{model-shrink-rate}: log-uniform distribution over $[10^{-5}, 10^{-2}]$ for StochasticRank;
    \item \textit{diffusion-temperature}: log-uniform distribution over $[10^8, 10^{11}]$ for StochasticRank;
    \item \textit{mu}: log-uniform distribution over $[10^{-2}, 10]$ for StochasticRank;
    \item \emph{num-leaves}: integer log-uniform distribution over $[16, 256]$ for LightGBM;
    \item \emph{min-data-in-leaf}: integer log-uniform distribution over $[1, 1000]$ for LightGBM; default value of $10$ for other algorithms.
\end{itemize}

The best hyper-parameters used for Table~\ref{tab:main_results} are listed in Tables~\ref{tab:params_ndcg}-\ref{tab:params_err}.

\section{Additional Experimental Details}\label{app:experiments}

\paragraph{Generalization gap}
Tables~\ref{tab:gen1}-\ref{tab:gen4} provide additional information on the generalization of different algorithms. The column ``train'' shows the corresponding error on the train set, while ``gap'' is the difference of the error between the train and test sets. A smaller value in ``gap'' indicates better generalization.

As expected, YetiLoss usually has larger generalization gaps than YetiRank (though there are some exceptions) since it is trained for a particular target quality function on the train set. Also, YetiLoss usually has better quality on the train set compared to YetiRank. Compared with LightGBM, all CatBoost-based algorithms usually have better generalization (see Table~\ref{tab:gen1}). We assume that this is because CatBoost uses oblivious decision trees known to be resistant to overfitting.

\paragraph{Number of neighbors}

Extended results on the analysis of the number of neighbors (including additional loss functions $\mathrm{NDCG}$@1 and $\mathrm{NDCG}$@5) can be found in Tables~\ref{tab:neighbors_first}-\ref{tab:neighbors_last}.

\paragraph{Stochastic smoothing}

Extended results on the analysis of stochastic smoothing (including additional loss functions $\mathrm{NDCG}$@1 and $\mathrm{NDCG}$@5) can be found in Tables~\ref{tab:noise_first}-\ref{tab:noise_last}.

\begin{table}
\caption{The effect of the number of neighbors for YetiLoss, $\mathrm{NDCG}$@1}
\label{tab:neighbors_first}
\centering
\begin{tabular}{l|ccc}
\toprule
 & Web & Yahoo & Istella \\
\midrule
1 & 48.24 & 74.64 & 70.36 \\
2 & 49.27 & 74.60 & 70.70 \\
all & 47.70 & 74.74 & 70.93 \\
\bottomrule
\end{tabular}
\end{table}
\begin{table}
\caption{The effect of the number of neighbors for YetiLoss, $\mathrm{NDCG}$@5}
\centering
\begin{tabular}{l|ccc}
\toprule
 & Web & Yahoo & Istella \\
\midrule
1 & 48.61 & 74.49 & 70.39 \\
2 & 49.02 & 74.29 & 70.61 \\
all & 48.36 & 74.65 & 70.44 \\
\bottomrule
\end{tabular}
\end{table}

\begin{table}
\caption{The effect of the number of neighbors for YetiLoss, $\mathrm{NDCG}$@10}
\centering
\begin{tabular}{l|ccc}
\toprule
 & Web & Yahoo & Istella \\
\midrule
1 & 50.76 & 77.77 & 76.32 \\
2 & 50.90 & 77.68 & 76.69 \\
3 & 50.99 & 77.76 & 76.74 \\
all & 50.56 & 77.90 & 76.79 \\
\bottomrule
\end{tabular}
\end{table}
\begin{table}
\caption{The effect of the number of neighbors for YetiLoss, $\mathrm{MRR}$}
\centering
\begin{tabular}{l|ccc}
\toprule
 & Web & Yahoo & Istella \\
\midrule
1 & 84.17 & 93.21 & 97.97 \\
2 & 84.18 & 93.75 & 97.99 \\
3 & 84.39 & 93.22 & 97.82 \\
all & 84.19 & 93.84 & 97.90 \\
\bottomrule
\end{tabular}
\end{table}

\begin{table}
\caption{The effect of the number of neighbors for YetiLoss, $\mathrm{MAP}$}
\centering
\begin{tabular}{l|ccc}
\toprule
 & Web & Yahoo & Istella \\
\midrule
1 & 62.62 & 89.37 & 91.07 \\
2 & 62.62 & 89.30 & 91.06 \\
3 & 62.54 & 89.38 & 91.16 \\
all & 62.32 & 89.36 & 90.43 \\
\bottomrule
\end{tabular}
\end{table}
\begin{table}
\caption{The effect of the number of neighbors for YetiLoss, $\mathrm{ERR}$}
\label{tab:neighbors_last}
\centering
\begin{tabular}{l|ccc}
\toprule
 & Web & Yahoo & Istella \\
\midrule
1 & 57.46 & 67.15 & 86.59 \\
2 & 57.91 & 67.28 & 86.71 \\
3 & 57.59 & 67.13 & 86.48 \\
all & 57.07 & 67.32 & 86.44 \\
\bottomrule
\end{tabular}
\end{table}



\begin{table}
\caption{The effect of stochastic smoothing for YetiRank, $\mathrm{NDCG}$@1}
\label{tab:noise_first}
\vspace{5pt}
\centering
\begin{tabular}{l|ccc}
\toprule
 & Web & Yahoo & Istella \\
\midrule
Logistic & 48.59 & 75.34 & 70.8 \\
Gaussian & 49.86 & 75.19 & 71.62 \\ 
No smoothing & 48.26 & 73.7 & 70.47 \\
\bottomrule
\end{tabular}

\end{table}

\begin{table}
\caption{The effect of stochastic smoothing for YetiRank, $\mathrm{NDCG}$@5}
\vspace{5pt}
\centering
\begin{tabular}{l|ccc}
\toprule
 & Web & Yahoo & Istella \\
\midrule
Logistic & 48.54 & 74.93 & 70.81 \\
Gaussuan & 49.4 & 74.99 & 70.94 \\
No smoothing & 47.52 & 72.14 & 69.82 \\
\bottomrule
\end{tabular}
\end{table}

\begin{table}
\caption{The effect of stochastic smoothing for YetiRank, $\mathrm{NDCG}$@10}
\vspace{5pt}
\centering
\begin{tabular}{l|ccc}
\toprule
 & Web & Yahoo & Istella \\
\midrule
Logistic & 50.75 & 77.99 & 77.15 \\
Gaussian & 51 & 78.07 & 77.23 \\
No smoothing & 49.05 & 75.57 & 75.8 \\
\bottomrule
\end{tabular}
\end{table}

\begin{table}
\caption{The effect of stochastic smoothing for YetiRank, $\mathrm{MRR}$}
\vspace{5pt}
\centering
\begin{tabular}{l|ccc}
\toprule
 & Web & Yahoo & Istella \\
\midrule
Logistic & 84.21 & 93.05 & 97.58 \\
Gaussian & 83.85 & 93.07 & 97.47 \\
No smoothing & 83.15 & 92.23 & 97.13 \\
\bottomrule
\end{tabular}
\end{table}

\begin{table}
\caption{The effect of stochastic smoothing for YetiRank, $\mathrm{MAP}$}
\vspace{5pt}
\centering
\begin{tabular}{l|ccc}
\toprule
 & Web & Yahoo & Istella \\
\midrule
Logistic & 62.07 & 88 & 89.09 \\
Gaussian & 61.85 & 87.8 & 89.02 \\
No smoothing & 60.38 & 86.85 & 87.47 \\
\bottomrule
\end{tabular}
\end{table}

\begin{table}
\caption{The effect of stochastic smoothing for YetiRank, $\mathrm{ERR}$}
\vspace{5pt}
\centering
\begin{tabular}{l|ccc}
\toprule
 & Web & Yahoo & Istella \\
\midrule
Logistic & 57.41 & 67.27 & 86.97 \\
Gaussian & 57.76 & 67.15 & 86.87 \\
\bottomrule
\end{tabular}
\end{table}

\newpage

\begin{table}
\caption{The effect of stochastic smoothing for YetiLoss, $\mathrm{NDCG}$@1}
\vspace{5pt}
\centering
\begin{tabular}{l|ccc}
\toprule
 & Web & Yahoo & Istella \\
\midrule
Logistic & 48.24 & 74.64 & 70.36 \\
Gaussian & 49.99 & 74.36 & 70.45 \\
No smoothing & 48.75 & 71.89 & 68.32 \\
\bottomrule
\end{tabular}

\end{table}

\begin{table}
\caption{The effect of stochastic smoothing for YetiLoss, $\mathrm{NDCG}$@5}
\vspace{5pt}
\centering
\begin{tabular}{l|ccc}
\toprule
 & Web & Yahoo & Istella \\
\midrule
Logistic & 48.61 & 74.49 & 70.39 \\
Gaussian & 49.25 & 74.17 & 70.37 \\
No smoothing & 46.85 & 71.51 & 68.46 \\
\bottomrule
\end{tabular}
\end{table}

\begin{table}
\caption{The effect of stochastic smoothing for YetiLoss, $\mathrm{NDCG}$@10}
\vspace{5pt}
\centering
\begin{tabular}{l|ccc}
\toprule
 & Web & Yahoo & Istella \\
\midrule
Logistic & 50.76 & 77.77 & 76.32 \\
Gaussian & 50.79 & 77.55 & 76.28 \\
No smoothing & 48.49 & 75.29 & 74.55 \\
\bottomrule
\end{tabular}
\end{table}

\begin{table}
\caption{The effect of stochastic smoothing for YetiLoss, $\mathrm{MRR}$}
\vspace{5pt}
\centering
\begin{tabular}{l|ccc}
\toprule
 & Web & Yahoo & Istella \\
\midrule
Logistic & 84.17 & 93.21 & 97.97 \\
Gaussian & 84.28 & 93.61 & 97.83 \\
No smoothing & 82.7 & 91.94 & 97.38 \\
\bottomrule
\end{tabular}
\end{table}

\begin{table}
\caption{The effect of stochastic smoothing for YetiLoss, $\mathrm{MAP}$}
\vspace{5pt}
\centering
\begin{tabular}{l|ccc}
\toprule
 & Web & Yahoo & Istella \\
\midrule
Logistic & 62.62 & 89.37 & 91.07 \\
Gaussian & 62.52 & 89.37 & 91.08 \\
No smoothing & 61.32 & 88.13 & 89.89 \\
\bottomrule
\end{tabular}
\end{table}

\begin{table}
\caption{The effect of stochastic smoothing for YetiLoss, $\mathrm{ERR}$}
\label{tab:noise_last}
\vspace{5pt}
\centering
\begin{tabular}{l|ccc}
\toprule
 & Web & Yahoo & Istella \\
\midrule
Logistic & 57.46 & 67.15 & 86.59 \\
Gaussian & 57.55 & 66.99 & 86.25 \\
\bottomrule
\end{tabular}
\end{table}

\begin{table*}[t]
\caption{Analysis of generalization gap for $\mathrm{NDCG}$@10}
\label{tab:gen1}
\vspace{5pt}
\begin{center}
\begin{small}
\begin{tabular}{lrrrrrrrrrrrr}
\toprule
 & \multicolumn{2}{c}{Web10K} & \multicolumn{2}{c}{Web30K} & \multicolumn{2}{c}{Yahoo S1} & \multicolumn{2}{c}{Yahoo S2} & \multicolumn{2}{c}{Istella} & \multicolumn{2}{c}{Istella-S} \\
Algorithm  & train & gap & train & gap & train & gap & train & gap & train & gap & train & gap \\
\midrule
YetiRank & 56.88 & 6.13 & 57.43 & 5.12 & 86.56 & 7.37 & 90.53 & 12.54 & 79.86 & 6.83 & 85.86 & 8.71 \\
YetiLoss & 57.68 & 6.92 & 52.48 & 3.57 & 80.46 & 5.34 & 94.83 & 17.06 & 77.03 & 5.57 & 85.23 & 8.91 \\
StochasticRank & 56.97 & 10.23 & 54.96 & 6.33 & 82.16 & 6.73 & 89.66 & 13.73 & 72.02 & 3.71 & 82.53 & 8.17 \\
LambdaMART & 55.25 & 5.87 & 55.72 & 4.79 & 83.35 & 4.72 & 90.48 & 13.17 & 73.38 & 3.55 & 80.12 & 4.57 \\
LightGBM & 56.04 & 9.25 & 59.58 & 10.37 & 84.99 & 9.67 & 84.79 & 9.39 & 82.36 & 9.61 & 87.73 & 11.77 \\
\bottomrule
\end{tabular}
\end{small}
\end{center}
\end{table*}

\begin{table*}[t]
\caption{Analysis of generalization gap for $\mathrm{MRR}$}
\label{tab:gen2}
\vskip 0.15in
\begin{center}
\begin{tabular}{lrrrrrrrrrrrr}
\toprule
 & \multicolumn{2}{c}{Web10K} & \multicolumn{2}{c}{Web30K} & \multicolumn{2}{c}{Yahoo S1} & \multicolumn{2}{c}{Yahoo S2} & \multicolumn{2}{c}{Istella} & \multicolumn{2}{c}{Istella-S} \\
Algorithm  & train & gap & train & gap & train & gap & train & gap & train & gap & train & gap \\
\midrule
YetiRank & 87.02 & 2.80 & 86.85 & 1.27 & 93.06 & 1.86 & 96.26 & 3.21 & 98.40 & 1.70 & 99.48 & 1.90 \\
YetiLoss & 87.82 & 3.64 & 87.64 & 1.63 & 94.93 & 4.04 & 97.88 & 4.67 & 98.98 & 1.82 & 99.85 & 1.88 \\
StochasticRank & 88.14 & 4.63 & 89.20 & 3.77 & 93.92 & 3.20 & 95.64 & 2.56 & 98.24 & 1.70 & 99.45 & 1.62 \\
LambdaMART & 86.11 & 3.52 & 85.27 & 1.15 & 93.26 & 2.51 & 96.01 & 3.67 & 97.60 & 2.42 & 99.66 & 2.70 \\
\bottomrule
\end{tabular}
\end{center}
\end{table*}

\begin{table*}[t]
\caption{Analysis of generalization gap for $\mathrm{MAP}$}
\label{tab:gen3}
\vskip 0.15in
\begin{center}
\begin{tabular}{lrrrrrrrrrrrr}
\toprule
 & \multicolumn{2}{c}{Web10K} & \multicolumn{2}{c}{Web30K} & \multicolumn{2}{c}{Yahoo S1} & \multicolumn{2}{c}{Yahoo S2} & \multicolumn{2}{c}{Istella} & \multicolumn{2}{c}{Istella-S} \\
Algorithm  & train & gap & train & gap & train & gap & train & gap & train & gap & train & gap \\
\midrule
YetiRank & 65.38 & 3.31 & 65.02 & 1.63 & 89.48 & 3.54 & 94.16 & 6.15 & 84.55 & 4.53 & 92.26 & 3.18 \\
YetiLoss & 65.86 & 3.24 & 66.22 & 2.20 & 90.70 & 4.24 & 96.06 & 6.69 & 89.40 & 6.74 & 96.27 & 5.20 \\
LambdaMART & 62.40 & 4.46 & 59.89 & 1.32 & 87.13 & 1.22 & 96.50 & 7.36 & 85.75 & 4.94 & 94.02 & 3.74 \\
\bottomrule
\end{tabular}
\end{center}
\end{table*}

\begin{table*}[t]
\caption{Analysis of generalization gap for $\mathrm{ERR}$}
\label{tab:gen4}
\vskip 0.15in
\begin{center}
\begin{tabular}{lrrrrrrrrrrrr}
\toprule
 & \multicolumn{2}{c}{Web10K} & \multicolumn{2}{c}{Web30K} & \multicolumn{2}{c}{Yahoo S1} & \multicolumn{2}{c}{Yahoo S2} & \multicolumn{2}{c}{Istella} & \multicolumn{2}{c}{Istella-S} \\
Algorithm  & train & gap & train & gap & train & gap & train & gap & train & gap & train & gap \\
\midrule
YetiRank & 61.80 & 4.39 & 62.53 & 3.00 & 70.34 & 3.64 & 73.40 & 6.13 & 91.30 & 4.02 & 94.53 & 7.55 \\
YetiLoss & 61.72 & 4.26 & 62.11 & 2.58 & 70.06 & 3.34 & 72.23 & 5.08 & 91.00 & 4.17 & 94.14 & 7.55 \\
StochasticRank & 62.82 & 5.88 & 62.81 & 3.48 & 70.00 & 3.36 & 73.26 & 6.09 & 87.77 & 2.33 & 91.82 & 5.84 \\
LambdaMART & 63.85 & 7.31 & 63.68 & 4.92 & 68.85 & 2.63 & 72.04 & 5.50 & 87.82 & 3.11 & 91.89 & 6.65 \\
\bottomrule
\end{tabular}
\end{center}
\end{table*}

\begin{table*}[t]
\caption{Chosen parameters for $\mathrm{NDCG}@10$}
\label{tab:params_ndcg}
\vskip 0.15in
\begin{center}
\begin{tabular}{llcccccc}
\toprule
 Algorithm & Parameter & Web10K & Web30K & Yahoo S1 & Yahoo S2 & Istella & Istella-S \\
\midrule
\multirow{3}{*}{YetiRank} &	depth & 8 & 8 & 8 & 7 & 8 & 8 \\
 & l2-leaf-reg & 0.03343 & 0.00100 & 0.00588 & 0.08913 & 0.00461 & 0.00100 \\ 
 & learning-rate & 0.05450 & 0.10085 & 0.10285 & 0.04172 & 0.18317 & 0.18166 \\
\midrule
\multirow{3}{*}{YetiLoss} &	depth & 8 & 8 & 8 & 8 & 8 & 8 \\
 & l2-leaf-reg & 0.00625 & 0.11205 & 0.10000 & 0.00313 & 0.13938 & 0.00100 \\ 
 & learning-rate & 0.05111 & 0.11108 & 0.09002 & 0.03479 & 0.25685 & 0.15083 \\
\midrule
\multirow{5}{*}{StochasticRank}	& depth & 10 & 10 & 10 & 8 & 10 & 10 \\
 & learning-rate & 0.18210 & 0.30549 & 0.12293 & 0.05506 & 1.00000 & 1.00000 \\
 & model-shrink-rate & 0.00001 & 0.00117 & 0.00049 & 0.01000 & 0.00075 & 0.00125 \\
 & diffusion-temperature & 675M & 63274M & 100M & 102M & 304M & 5873M \\ 
 & mu & 0.02034 & 0.01314 & 0.06495 & 0.04142 & 0.01190 & 0.01794 \\
\midrule
\multirow{3}{*}{LambdaMART} & depth & 8 & 8 & 8 & 6 & 8 & 7 \\
 & l2-leaf-reg & 0.00394 & 0.00170 & 0.00176 & 0.00100 & 0.00100 & 0.00100 \\ 
 & learning-rate & 0.07478 & 0.14310 & 0.14611 & 0.10625 & 0.18117 & 0.15533 \\
\midrule
\multirow{3}{*}{LightGBM} & depth & 14 & 14 & 16 & 9 & 14 & 16 \\
 & learning-rate & 0.16656 & 0.12405 & 0.09355 & 0.12593 & 0.22409 & 0.22599 \\
 & num-leaves & 97 & 256 & 256 & 16 & 256 & 256 \\
 & min-data-in-leaf & 1 & 1 & 2 & 11 & 9 & 1 \\
 & l2-leaf-reg & 73.06401 & 0.39895 & 6.58340 & 11.54666 & 0.74747 & 1.76279 \\
\bottomrule
\end{tabular}
\end{center}
\end{table*}

\begin{table*}[t]
\caption{Chosen parameters for $\mathrm{MRR}$}
\label{tab:mrr}
\vskip 0.15in
\begin{center}
\begin{tabular}{llcccccc}
\toprule
 Algorithm & Parameter & Web10K & Web30K & Yahoo S1 & Yahoo S2 & Istella & Istella-S \\
\midrule
\multirow{3}{*}{YetiRank} &	depth & 7 & 7 & 8 & 6 & 8 & 8 \\
 & l2-leaf-reg & 0.23902 & 0.55462 & 0.01642 & 6.24819 & 0.04491 & 0.11146 \\ 
 & learning-rate & 0.09387 & 0.22263 & 0.07333 & 0.23151 & 0.14079 & 0.17564 \\
\midrule
\multirow{3}{*}{YetiLoss} &	depth & 6 & 8 & 7 & 8 & 8 & 8 \\
 & l2-leaf-reg & 0.71013 & 0.32674 & 0.10000 & 0.02186 & 0.10000 & 0.00356 \\ 
 & learning-rate & 0.16364 & 0.07572 & 0.15166 & 0.07400 & 0.18719 & 0.06319 \\
\midrule
\multirow{5}{*}{StochasticRank}	& depth & 7 & 10 & 10 & 6 & 10 & 10 \\
 & learning-rate & 0.19705 & 0.34981 & 0.14603 & 0.04305 & 1.00000 & 0.36481 \\
 & model-shrink-rate & 0.00001 & 0.00016 & 0.01000 & 0.00001 & 0.00006 & 0.00031 \\
 & diffusion-temperature & 251M & 8890M & 100M & 100M & 100M & 984M \\
 & mu & 0.01120 & 0.01000 & 0.03325 & 0.01689 & 0.03698 & 0.01000 \\
\midrule
\multirow{3}{*}{LambdaMART} & depth & 6 & 6 & 8 & 8 & 6 & 6 \\
 & l2-leaf-reg & 0.00769 & 0.04354 & 0.00100 & 0.01220 & 0.00100 & 0.00100 \\ 
 & learning-rate & 0.23696 & 0.08981 & 0.07174 & 0.12050 & 0.09162 & 0.09379 \\
\bottomrule
\end{tabular}
\end{center}
\end{table*}

\begin{table*}[t]
\caption{Chosen parameters for $\mathrm{MAP}$}
\label{tab:params_map}
\vskip 0.15in
\begin{center}
\begin{tabular}{llcccccc}
\toprule
 Algorithm & Parameter & Web10K & Web30K & Yahoo S1 & Yahoo S2 & Istella & Istella-S \\
\midrule
\multirow{3}{*}{YetiRank} &	depth & 8 & 8 & 8 & 7 & 8 & 8 \\
 & l2-leaf-reg & 0.01157 & 0.00320 & 0.00708 & 0.20218 & 0.00287 & 0.02457 \\ 
 & learning-rate & 0.05033 & 0.09699 & 0.11392 & 0.05680 & 0.18562 & 0.15196 \\
\midrule
\multirow{3}{*}{YetiLoss} &	depth & 8 & 8 & 8 & 8 & 8 & 8 \\
 & l2-leaf-reg & 0.54619 & 0.04957 & 0.05314 & 0.00274 & 0.00819 & 0.02790 \\ 
 & learning-rate & 0.05031 & 0.07755 & 0.06111 & 0.01103 & 0.25184 & 0.25536 \\
\midrule
\multirow{3}{*}{LambdaMART} & depth & 8 & 8 & 8 & 7 & 8 & 8 \\
 & l2-leaf-reg & 0.01119 & 0.00441 & 0.00100 & 0.00879 & 0.00100 & 0.00100 \\ 
 & learning-rate & 0.01769 & 0.01267 & 0.02798 & 0.04096 & 0.18745 & 0.16072 \\
\bottomrule
\end{tabular}
\end{center}
\end{table*}

\begin{table*}[t]
\caption{Chosen parameters for $\mathrm{ERR}$}
\label{tab:params_err}
\vskip 0.15in
\begin{center}
\begin{tabular}{llcccccc}
\toprule
 Algorithm & Parameter & Web10K & Web30K & Yahoo S1 & Yahoo S2 & Istella & Istella-S \\
\midrule
\multirow{3}{*}{YetiRank} &	depth & 8 & 8 & 8 & 8 & 8 & 8 \\
 & l2-leaf-reg & 0.06115 & 0.00100 & 0.00309 & 0.01318 & 0.02537 & 0.00100 \\ 
 & learning-rate & 0.06827 & 0.09341 & 0.09871 & 0.02678 & 0.17561 & 0.16226 \\
\midrule
\multirow{3}{*}{YetiLoss} &	depth & 8 & 8 & 8 & 6 & 8 & 8 \\
 & l2-leaf-reg & 0.01401 & 0.00175 & 0.00966 & 0.05289 & 0.00100 & 0.00160 \\ 
 & learning-rate & 0.04484 & 0.07811 & 0.06638 & 0.05021 & 0.12214 & 0.09334 \\
\midrule
\multirow{5}{*}{StochasticRank}	& depth & 10 & 10 & 10 & 10 & 10 & 10 \\
 & learning-rate & 0.14777 & 0.30041 & 0.25101 & 0.05651 & 0.54801 & 0.39642 \\
 & model-shrink-rate & 0.00042 & 0.00001 & 0.00004 & 0.01000 & 0.00085 & 0.00206 \\
 & diffusion-temperature & 100M & 589M & 344M & 1425M & 100B & 100M \\
 & mu & 0.12610 & 0.02085 & 0.12851 & 0.18189 & 0.01375 & 0.01000 \\
\midrule
\multirow{3}{*}{LambdaMART} & depth & 8 & 8 & 8 & 6 & 6 & 6 \\
 & l2-leaf-reg & 0.00100 & 0.00100 & 0.00116 & 0.00100 & 0.00100 & 0.00100 \\ 
 & learning-rate & 0.05517 & 0.10631 & 0.09575 & 0.04895 & 0.15040 & 0.15975 \\
\bottomrule
\end{tabular}
\end{center}
\end{table*}

\end{document}